\documentclass{article}

\usepackage{microtype}
\usepackage{graphicx}
\usepackage{subcaption}
\usepackage{booktabs} 

\usepackage{hyperref}

\usepackage[accepted]{icml2026}

\usepackage{amsmath}
\usepackage{amssymb}
\usepackage{mathtools}
\usepackage{amsthm}

\usepackage{booktabs}
\usepackage{multirow}
\usepackage{array}
\usepackage{tabularx}
\usepackage{threeparttable}

\renewcommand{\arraystretch}{1.6}

\definecolor{gaincolor}{RGB}{0,120,0}
\definecolor{dropcolor}{RGB}{180,0,0}

\usepackage[capitalize,noabbrev]{cleveref}

\theoremstyle{plain}

\theoremstyle{definition}

\theoremstyle{remark}

\usepackage[textsize=tiny]{todonotes}

\icmltitlerunning{Titans-as-a-Layer: Test-Time Memory for Conversational Speech Emotion Recognition}

\begin{document}

\twocolumn[
  \icmltitle{Titans-as-a-Layer: Test-Time Memory for Conversational Speech Emotion Recognition}



  \icmlsetsymbol{equal}{*}

  \begin{icmlauthorlist}
    \icmlauthor{Daniel Chen}{uoa}
    \icmlauthor{Qicong Hu}{uoa}
    \icmlauthor{Yang Xiao}{uom}
    \icmlauthor{Ting Dang}{uom}
    \icmlauthor{Hong Jia}{uoa}
  \end{icmlauthorlist}

  \icmlaffiliation{uoa}{Department of Computer Science, University of Auckland, Auckland, New Zealand}
  \icmlaffiliation{uom}{School of Computer and Information Technology, Melbourne, Australia}

  \icmlcorrespondingauthor{Ting Dang}{ting.dang@unimelb.edu.au}

  \icmlkeywords{Titans, Audio Large Language Models, Memory, Speech Emotion Recognition}

  \vskip 0.3in
]




\printAffiliationsAndNotice{}

\begin{abstract}
Speech emotion recognition (SER) is commonly formulated as utterance-level classification, although conversational emotion depends on a speaker's usual vocal range and the emotional context established by previous utterances. Speech-language models provide strong pretrained acoustic and semantic representations, and can adapts them to SER labels via finetune, but this mechanism still missing per-dialogue state. We study whether test-time neural memory can supply this missing context while leaving the large audio language models (LALMs) backbone intact. Building on Titans, we introduce a plug-and-play Memory-as-a-Layer (MAL) adapter that writes dialogue history into a small neural memory and reads it back as an audio-token-aligned residual update, avoiding changes to the host model's token positions. Across different audio LLMs and emotion recognition datasets evaluations, our design improves SER performs across different evaluation metrics, supporting test-time memory as a residual contextual mechanism for conversational SER.
\end{abstract}

\section{Introduction}
Speech emotion is conveyed not only by lexical content, but also by the manner in which that content is produced. In conversational speech emotion recognition (SER), the acoustic evidence supporting an emotion label is therefore rarely contained within an isolated utterance. Prosodic and paralinguistic cues, including pitch, intensity, speaking rate and voice quality, must be interpreted relative to the speaker’s \textit{recent vocal behaviour} and the \textit{affective evolution of the dialogue}. We refer to this short-term vocal reference as the speaker baseline. For instance, an increase in pitch may signal anger for one speaker, enthusiasm for another, or no salient affective change if it lies within that speaker’s recent expressive range. Conversational SER is thus inherently both local and contextual: it depends on the current waveform while also requiring information from preceding utterances in the same interaction.

Large-scale speech pretraining and audio–text alignment provide a strong foundation for SER, as contemporary speech-language models encode both linguistic content and non-lexical attributes such as prosody, energy and speaking style~\cite{radford2023whisper,tang2024salmonn,chu2024qwen2audio,kong2024audioflamingo}. In standard SER pipelines, however, these representations are commonly used at the utterance level: each speech segment is encoded and classified independently~\cite{jia2025beyond,zhang2025menta}. Consequently, evidence that is distributed across speakers and dialogue history cannot be retained intrinsically, and must instead be introduced through explicit context tokens, concatenated inputs, or an additional mechanism for maintaining dialogue state.

LoRA~\cite{hu2022lora} efficiently adapts LALMs to SER without full-model fine-tuning, but the learned update is static and dialogue-independent. It improves task alignment, yet cannot maintain speaker baselines, affective trajectories, or previous-utterance information. Extending the attention context may also disturb the positional structure learned during pretraining. These limitations motivate inference-time memory with a fixed backbone. Titans~\cite{behrouz2024titans} treats memory as a compact neural module updated during inference, storing prior information as model state rather than explicit context tokens. For conversational SER, this separates task adaptation from dialogue-state modelling: LoRA aligns the model to emotion recognition, while test-time memory preserves dialogue-specific evidence for subsequent utterances.

Motivated by these, we introduce a plug-and-play \textit{Memory-as-a-Layer} (MAL) adapter based on Titans for LALMs. MAL inserts a Titans-style neural memory branch within the language-model stack and injects its output as an audio-token-aligned residual update. The adapter does not alter the pretrained LALMs: it neither prepends memory tokens to the LALMs input, nor shifts audio-token positions, nor requires the backbone to process the dialogue history as an extended context window. Instead, dialogue information is written into and retrieved from a separate memory state. Across multiple datasets, LALMs backbones and evaluation metrics, MAL consistently improves conversational SER performance, demonstrating that test-time memory provides a robust mechanism for incorporating dialogue-level context without compromising the pretrained model interface.

Our contributions are:
\begin{itemize}
    \item We introduce a Titans-style memory branch that updates audio-token representations in place, preserving the host model's token layout while adding an explicit pathway for dialogue-level state.
    \item We introduce a two-stage optimisation scheme in which LoRA first provides task adaptation for SER, after which the LoRA-tuned model is frozen and MAL is trained to supply dialogue-specific contextual refinement. This design encourages memory to act as an additive state mechanism rather than as a competing task adapter.
    \item We evaluate multiple LALMs across diverse SER datasets and metrics, demonstrating that the proposed method achieves state-of-the-art SER performance.
\end{itemize}
    
\section{Related Work}
\subsection{Memory in large language models} 

Existing memory mechanisms for sequence models address the limited context available to standard Transformers. Transformer-XL introduces segment-level recurrence~\cite{dai2019transformerxl}; Compressive Transformer stores compressed representations of past activations~\cite{rae2020compressive}; Recurrent Memory Transformer uses learned memory tokens~\cite{bulatov2022rmt}; and linear-attention variants improve the efficiency of long-range sequence modeling~\cite{yang2024gla}. Titans~\cite{behrouz2024titans} extends this direction by representing long-term memory as a small neural module whose parameters are updated during inference, together with sliding-window attention and learned persistent memory tokens. Titans-style memory has also been adapted beyond text: VideoTitan~\cite{park2025videotitans} applies the mechanism to video sequence modeling and reports strong results on long video benchmarks, including WeatherBench~\cite{rasp2020weatherbench}. However, these approaches do not directly address frozen speech-language models for conversational SER, where the memory mechanism must preserve the audio-token interface while carrying information across utterances.

\subsection{Static task adapters and adaptive memory.}
Adapter-based fine-tuning methods such as LoRA~\cite{hu2022lora} adapt pretrained models by adding trainable low-rank updates while keeping the backbone fixed. For SER, such adapters can learn a task-specific mapping from pretrained speech representations to emotion labels with far fewer trainable parameters than full fine-tuning. This makes LoRA an appropriate task-adaptation baseline for large speech-language models. However, a trained adapter is static at test time: the same learned update is applied to every dialogue, regardless of the speaker's vocal range or the preceding emotional context. This leaves open the question of whether an adaptive memory component can add dialogue-specific state after task adaptation has been learned.

\section{Method}
\label{sec:method}

\subsection{Test-time memory}

We consider conversational speech emotion recognition, where a dialogue is an
ordered utterance stream $\mathcal{D}=(x_1,\ldots,x_N)$ with utterance-level
labels $(y_1,\ldots,y_N)$. Each utterance $x_i$ is encoded into $T_i$
audio-token embeddings and inserted into the language-model sequence at the
audio-token positions.

We use \textit{test-time memory} to denote a dialogue-specific state that is
updated online during inference while the backbone and trained adapters remain
fixed. For utterance $x_i$, the model predicts using the current utterance and
the memory accumulated from previous utterances:
\begin{equation}
    \hat{y}_i = f_{\Theta}(x_i, \mathcal{S}_{i-1}),
    \qquad
    \mathcal{S}_i = U_{\Phi}(\mathcal{S}_{i-1}, x_i).
    \label{eq:test-time-memory}
\end{equation}
Here $\mathcal{S}_i$ is the dialogue memory state after utterance $x_i$,
$f_{\Theta}$ is the fixed inference model, and $U_{\Phi}$ is the learned memory
update rule. The memory state is reset at dialogue boundaries. At test time, no
model parameters are optimized; only the dialogue-specific state
$\mathcal{S}_i$ evolves.

For MAL, the memory state is layer-specific:
\begin{equation}
    \mathcal{S}_i = \{S_i^{(\ell)}\}_{\ell=0}^{L-1},
\end{equation}
where $S_i^{(\ell)}$ is the NeuralMemory state for language-model block $\ell$
after processing utterance $x_i$.

\subsection{Memory integration ablation} 

We compare three ways of integrating a Titans memory branch into a frozen LALM. This ablation uses Ultravox~v0.4 on IEMOCAP and attaches a single memory branch, so that the comparison isolates the injection mechanism. Let $h \in \mathbb{R}^{T \times D}$ be the audio-token hidden states in the language-model residual stream, let $P \in \mathbb{R}^{N_p \times D}$ be persistent memory tokens, and let $\mathcal{M}$ be a Titans NeuralMemory module. We write $[\,\cdot\,;\,\cdot\,]$ for sequence concatenation and $[N_p{+}1{:}]$ for removing the persistent-token prefix. The audio-aligned memory output is
\begin{equation}
    m = \mathcal{M}([P;h])[N_p{+}1{:}],
    \qquad
    m \in \mathbb{R}^{T \times D}.
\end{equation}

\textbf{Memory as Context (MAC)} prepends memory to the language-model input:
\begin{equation}
    H_{\mathrm{MAC}} = [\,P \;;\; m \;;\; h\,].
\end{equation}
This increases the sequence length and shifts the positional indices of the
original tokens.

\textbf{Memory as Gating (MAG)} keeps the sequence length fixed and gates the residual stream:
\begin{equation}
    H_{\mathrm{MAG}} = h \odot \sigma(W_g m),
\end{equation}
where $W_g$ is learned, $\sigma(\cdot)$ is the sigmoid function, and $\odot$ denotes elementwise multiplication.

\textbf{Memory-as-a-Layer (MAL)} instead applies an additive memory update to the audio-token hidden states while preserving the original token positions. The zero-shot baseline obtains 44.52 WF1. MAC improves this to 49.90 WF1, MAG obtains 50.32 WF1, and MAL reaches 57.48 WF1. Therefore, we use MAL as the main memory integration strategy.

\begin{table*}[t]
\centering
\caption{Model Performance Results Across Benchmarks. \textbf{Bold} = best and \underline{underline} = second-best within each fully evaluated backbone--dataset block.}
\label{tab:titan_results}
\scriptsize
\setlength{\tabcolsep}{5.5pt}
\renewcommand{\arraystretch}{1.18}
\begin{threeparttable}
\resizebox{\textwidth}{!}{%
\begin{tabular}{@{}lccccccc@{}}
\toprule
\multirow{2}{*}{\textbf{Model}}
  & \multicolumn{3}{c}{\textbf{IEMOCAP}}
  & \multicolumn{2}{c}{\textbf{MELD}}
  & \multicolumn{2}{c}{\textbf{Multidialog}} \\
\cmidrule(lr){2-4} \cmidrule(lr){5-6} \cmidrule(lr){7-8}
  & \textbf{WA (\%)} & \textbf{UA (\%)} & \textbf{WF1 (\%)}
  & \textbf{WF1 (\%)} & \textbf{Macro-F1 (\%)}
  & \textbf{WF1 (\%)} & \textbf{Macro-F1 (\%)} \\
\midrule
 
Qwen2-Audio         & 58.92 & 59.82 & 58.25 & 39.05 & 20.56 & 28.27 &  17.65\\
\quad + LoRA             & \underline{82.40} & \underline{82.61} & \underline{82.42} & \underline{54.45} & \underline{38.32} & \underline{56.37} & \underline{37.16} \\
\quad + Titans+LoRA     & \textbf{83.66} & \textbf{84.20} & \textbf{83.67} & \textbf{56.84} & \textbf{40.99} & \textbf{57.54}  &  \textbf{37.94} \\
\midrule
 
Audio Flamingo 3          & 76.57 & 76.45 & 76.31 & 45.14 & 30.11 & 31.64  & 21.08 \\
\quad + LoRA               & \underline{84.59} & \underline{85.02} & \underline{84.57} & \underline{57.81} & \underline{44.43} & \underline{55.11} & \underline{34.78} \\
\quad + Titans+LoRA       & \textbf{85.21} & \textbf{85.37} & \textbf{85.20} & \textbf{58.18} & \textbf{44.66} & \textbf{56.07} & \textbf{35.82} \\
\midrule
 
Ultravox-v0.4    & 45.29 & 47.09 & 44.53 & 25.33 & 22.70 & 34.24 & 19.78 \\
\quad + LoRA         & \underline{63.89} & \underline{62.84} & \underline{63.78} & \underline{48.58} & \underline{31.97} & \underline{54.85} & \underline{30.74} \\
\quad + Titans+LoRA & \textbf{65.49} & \textbf{65.43} & \textbf{65.61} & \textbf{49.63} & \textbf{32.64} & \textbf{55.04}  & \textbf{33.32} \\
 
\bottomrule
\end{tabular}}
\begin{tablenotes}
  \scriptsize
  \item WA\,=\,Weighted Accuracy; UA\,=\,Unweighted Accuracy; WF1\,=\,Weighted F1.
\end{tablenotes}
\end{threeparttable}
\vspace{-1.5em}
\end{table*}

\subsection{Memory-as-a-Layer (MAL) for LALM}

For utterance $x_i$ and language-model block $\ell$, let
$h_{i,\ell} \in \mathbb{R}^{T_i \times D}$ denote the hidden states at the audio-token positions before the block. MAL modifies these states in place: it does not add, remove, or reorder tokens in the host sequence. The memory branch uses an internal dimension $d_m$, so the audio states are projected into memory space and projected back afterward.

For each block $\ell$, MAL computes
\begin{align}
    z_{i,\ell}
    &=
    \big[\, P_\ell \;;\; W_{\mathrm{in}}^{(\ell)} h_{i,\ell} \,\big], \\
    (r_{i,\ell}, S_i^{(\ell)})
    &=
    \mathcal{M}_\ell\!\left(z_{i,\ell}; S_{i-1}^{(\ell)}\right), \\
    \delta_{i,\ell}
    &=
    W_{\mathrm{out}}^{(\ell)} r_{i,\ell}[N_p{+}1{:}], \\
    \tilde{h}_{i,\ell}
    &=
    h_{i,\ell} + \tanh(\alpha_\ell)\,\delta_{i,\ell}.
\end{align}
Here $W_{\mathrm{in}}^{(\ell)}{:}\;D{\to}d_m$ and $W_{\mathrm{out}}^{(\ell)}{:}\;d_m{\to}D$ are learned projections, $P_\ell \in \mathbb{R}^{N_p \times d_m}$ are persistent memory tokens, and $\mathcal{M}_\ell$ is a Titans NeuralMemory module~\cite{behrouz2024titans}. The module reads the previous dialogue state $S_{i-1}^{(\ell)}$, processes the current memory input $z_{i,\ell}$, returns memory output $r_{i,\ell}$, and updates the state to $S_i^{(\ell)}$ for subsequent utterances.

After the memory branch, we discard the $N_p$ persistent-token prefix and keep only the $T_i$ audio-aligned outputs. The scalar gate $\alpha_\ell \in \mathbb{R}$ is initialized to zero, so $\tanh(\alpha_\ell)=0$ at initialization and
$\tilde{h}_{i,\ell}=h_{i,\ell}$. Thus, adding MAL initially preserves the frozen host computation.

We attach independent MAL branches to every language-model layer. In the placement ablation, dense every-layer placement with smaller branch capacity ($d_m{=}128$, $N_p{=}8$) outperformed sparser placements, such as every-$8$th layer at quartile positions $\{7,15,23,31\}$ with $d_m{=}256$.

\paragraph{Trainable parameters.}
The trainable MAL parameters are
\begin{equation}
    \{W_{\mathrm{in}}^{(\ell)}, W_{\mathrm{out}}^{(\ell)},
    \mathcal{M}_\ell, \alpha_\ell, P_\ell\}_{\ell=0}^{L-1}.
\end{equation}
All backbone, audio-encoder, and projector parameters remain frozen.

\begin{figure}[t]
    \centering
    \includegraphics[width=0.75\linewidth]{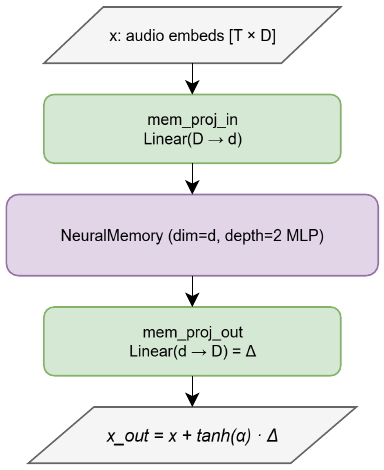}
    \caption{Memory-as-a-Layer branch architecture. Audio embeddings are projected down to the memory dimension $d$, passed through a Titans NeuralMemory module (depth-2 MLP), projected back to $D$, and added to the original embeddings through a zero-initialised residual gate $\tilde{h} = h + \tanh(\alpha) \cdot \delta$.}
    \label{fig:mal-branch}
    \vspace{-1.5em}
\end{figure}

\subsection{Sequential training schedule}
\label{sec:schedule}

We use a sequential schedule to separate static task adaptation from dialogue-state memory learning. Jointly optimizing LoRA and MAL did not improve over the LoRA-only baseline, likely because both modules modify the same residual stream while only MAL maintains state across utterances. Diagnostic ablations showed that removing LoRA after joint training had little effect, whereas removing Titans memory substantially degraded performance. We therefore train in two phases.

\paragraph{Phase 1: task adaptation.} We train LoRA with MAL removed, so the model learns the SER decision boundary without memory updates. LoRA is trained for three epochs with rank $r{=}32$, scaling $\alpha_{\mathrm{LoRA}}{=}64$, AdamW, and learning rate $2{\times}10^{-5}$.

\paragraph{Phase 2: memory training.} We select the best LoRA checkpoint, then freeze LoRA, the LALM backbone, audio encoder, and projector. MAL is trained for two epochs while all other components remain fixed. The projection and NeuralMemory parameters $\{W_{\mathrm{in}}, W_{\mathrm{out}}, \mathcal{M}\}$ use learning rate $1{\times}10^{-4}$ and gradient clipping threshold $1.0$; the residual gates $\{\alpha_\ell\}$ use learning rate $2{\times}10^{-4}$ and clipping threshold $2.0$.

\section{Experimental setup}
\subsection{Datasets}

We use three conversational speech emotion recognition datasets for the evaluated settings. For IEMOCAP~\cite{busso2008iemocap}, we used its standard four-class configuration (angry/happy/neutral/sad), with $5{,}531$ utterances across $5$ sessions and two consistent speakers per session, evaluated under leave-one-session-out (LOSO) cross-validation.
For MELD~\cite{poria2019meld}, we tested on the seven-class multi-party TV-dialogue setting with the standard train/dev/test split, totalling 13,706 utterances. For MultiDialog~\cite{park2024multidialog}, we used the gold subset of $9$ emotion-accurate actors, containing 6,681 dialogues and 47,004 utterances with the same seven-class schema as MELD.

\subsection{Models}

We use three models to assess whether the memory effect persists across different backbone designs: Ultravox~v0.4~\cite{ultravox2024}, Qwen2-Audio~\cite{chu2024qwen2audio}, and Audio Flamingo~3~\cite{goel2025audioflamingo3}. 




\section{Results}



The results in Table~\ref{tab:titan_results} show that incorporating Titans consistently improves performance over both the original backbone models and standard LoRA fine-tuning across the fully evaluated settings. On IEMOCAP, Titans+LoRA achieves the best results for all three backbones, improving over LoRA by up to 1.60\% in WA, 2.59\% in UA, and 1.83\% in WF1. These gains are particularly evident for Ultravox-v0.4, where Titans+LoRA increases WF1 from 63.78\% to 65.61\%, indicating that Titans provides additional modelling benefits beyond parameter-efficient fine-tuning alone. Similar improvements are observed on MELD, where Titans+LoRA improves Audio Flamingo 3 from 57.81\% to 58.18\% WF1 and from 44.43\% to 44.66\% Macro-F1, while also improving Ultravox-v0.4 from 48.58\% to 49.63\% WF1 and from 31.97\% to 32.64\% Macro-F1. On Multidialog, Titans+LoRA further improves Ultravox-v0.4, especially in Macro-F1, increasing performance from 30.74\% to 33.32\%. This suggests that Titans is particularly effective in improving recognition of underrepresented classes. Overall, the consistent gains across datasets, metrics, and backbone architectures demonstrate that Titans complements LoRA by enhancing the model's ability to capture task-relevant emotional and conversational representations.

\section{Conclusion}

Through this work, we study whether a NeuralMemory module (Titans) can augment frozen, LoRA-tuned LALMs with cross-utterance memory at test time. The evaluated settings show additive gains over the corresponding LoRA baselines, with the clearest evidence on IEMOCAP and additional support from the evaluated MELD runs. The key finding is that a sequential training schedule improves memory in a LoRA-tuned model for speech emotion recognition. These results support test-time memory as a lightweight complement to parameter-efficient fine-tuning for conversational emotion
recognition, while broader coverage across all backbone--dataset combinations and comparisons against other memory architectures remain important next steps.




\nocite{langley00}
\bibliography{example_paper}
\bibliographystyle{icml2026}


\end{document}